\documentclass[pdflatex,sn-mathphys-num]{sn-jnl}

\usepackage{graphicx}
\usepackage{subcaption}
\usepackage{multirow}
\usepackage{amsmath,amssymb,amsfonts}
\usepackage{amsthm}
\usepackage{mathrsfs}
\usepackage[title]{appendix}
\usepackage{xcolor}
\usepackage{textcomp}
\usepackage{manyfoot}
\usepackage{booktabs}
\usepackage{algorithm}
\usepackage{algorithmicx}
\usepackage{algpseudocode}
\usepackage{listings}
\usepackage{longtable}
\usepackage{booktabs} 
\usepackage{makecell}

\makeatletter
\renewcommand{\equalcont}[1]{}
\makeatother

\theoremstyle{thmstyleone}

\theoremstyle{thmstyletwo}

\theoremstyle{thmstylethree}

\begin{document}

\title{Energy Landscapes Enable Reliable Abstention in Retrieval-Augmented Large Language Models for Healthcare}

\author[1]{\fnm{Ravi} \sur{Shankar}}
\author[1]{\fnm{Sheng} \sur{Wong}}
\author[2]{\fnm{Lin} \sur{Li}}
\author[3]{\fnm{Magdalena} \sur{Bachmann}}
\author[3]{\fnm{Alex} \sur{Silverthorne}}
\author[1]{\fnm{Beth} \sur{Albert}}
\author*[1]{\fnm{Gabriel Davis} \sur{Jones}}\email{gabriel.jones@wrh.ox.ac.uk}

\affil[1]{\orgdiv{Oxford Digital Health Labs, Nuffield Department of Women's and Reproductive Health}, \orgname{University of Oxford}, \orgaddress{\city{Oxford}, \country{UK}}}
\affil[2]{\orgdiv{OATML, Department of Computer Science}, \orgname{University of Oxford}, \orgaddress{\city{Oxford}, \country{UK}}}
\affil[3]{\orgdiv{Nuffield Department of Women's and Reproductive Health}, \orgname{University of Oxford}, \orgaddress{\city{Oxford}, \country{UK}}}

\abstract{Reliable abstention is critical for retrieval-augmented generation (RAG) systems, particularly in safety-critical domains such as women’s health, where incorrect answers can lead to harm. We present an energy-based model (EBM) that learns a smooth energy landscape over a dense semantic corpus of 2.6M guideline-derived questions, enabling the system to decide when to generate or abstain. We benchmark the EBM against a calibrated softmax baseline and a k-nearest neighbour (kNN) density heuristic across both easy and hard abstention splits, where hard cases are semantically challenging near-distribution queries. The EBM achieves superior abstention performance abstention on semantically hard cases, reaching AUROC 0.961 versus 0.950 for softmax, while also reducing FPR@95 (0.235 vs 0.331). On easy negatives, performance is comparable across methods, but the EBM’s advantage becomes most pronounced in safety-critical hard distributions. A comprehensive ablation with controlled negative sampling and fair data exposure shows that robustness stems primarily from the energy scoring head, while the inclusion or exclusion of specific negative types (hard, easy, mixed) sharpens decision boundaries but is not essential for generalisation to hard cases. These results demonstrate that energy-based abstention scoring offers a more reliable confidence signal than probability-based softmax confidence, providing a scalable and interpretable foundation for safe RAG systems.}


\maketitle

\section{Introduction}\label{sec:introduction}

Large language models (LLMs) coupled with retrieval-augmented generation (RAG) 
\citep{lewis2020rag} are being piloted for clinical decision support 
\citep{thirunavukarasu2023llmmed,nori2023gpt4medical}. 
They can synthesise guidance across large corpora, yet they also generate fluent errors 
when inputs fall outside scope or when retrieved evidence is sparse or misleading. 
In safety-critical care, such failures erode trust and can cause harm. 
Robust abstention is therefore a first-order requirement 
\citep{chow1970optimum,geifman2017selective,liu2020energy}: 
the system should recognise when not to answer and instead expand retrieval, escalate, or defer to a human expert.

RAG is particularly promising where the knowledge corpus can be strictly controlled. 
In healthcare, evidence-based guidelines, drug formularies, and institution-specific protocols can be curated, versioned, and indexed so that generation is constrained to cited sources with known provenance and update cycles. 
This limits reliance on memorised web text, reduces parametric drift, enables auditability, and supports time-bounded answers. 
In women’s health, for example, authoritative guidance from bodies such as RCOG, NICE, and WHO can anchor responses to accepted standards of care. 
Nevertheless, reliability depends on recall and scope: if relevant material is not retrieved or the query falls outside the indexed corpus, the model may generalise beyond evidential support. 
In these settings, abstention and explicit self-assessment of evidential sufficiency are critical.

In healthcare applications, two distinct classes of query should trigger abstention: 
(i) domain-irrelevant prompts that fall outside healthcare expertise, for example questions related to finance or economics, where the correct behaviour is immediate redirection; and 
(ii) domain-relevant but out-of-scope queries relative to the model's training or validation, such as applying pregnancy-specific diabetes or hypertension protocols to non-pregnant adults, paediatric gynaecology when training covered only adult care, oncological queries requiring therapies, dosing, or trial evidence not present in the corpus, site-specific policies absent from the training data, or modality shifts such as image-first triage when the model was trained only on text. 
These near-distribution queries are hazardous because they are semantically close to in-scope content and can elicit persuasive but unsafe answers. 
Abstention should be the default unless retrieved evidence is sufficiently specific and in-distribution.

Prior work in women’s health underscores the need for principled abstention. In a head-to-head evaluation with questions from the UK Royal College of Obstetricians and Gynaecologists, ChatGPT achieved moderate accuracy on basic science but only coin-toss performance on clinical reasoning, while often expressing high confidence irrespective of correctness, indicating unreliable self-assessment \citep{Bachmann2024WomenHealthLLM}. Meaning-level uncertainty signals (for example semantic entropy) offer a more discriminative path: \emph{semantic entropy} outperformed perplexity for identifying unreliable outputs on obstetrics and gynaecology questions, and achieved expert-validated discrimination approaching ceiling, supporting the value of pre-generation uncertainty checks and deferral mechanisms \citep{PennyDimri2025SemanticEntropy}. However, computing semantic entropy typically requires first generating multiple candidate responses to estimate the meaning distribution, which adds latency and compute cost.

Foundational work on abstention frames the problem as a reject option in statistical decision theory, and selective prediction formalises the coverage–risk trade-off with standard risk–coverage evaluation \citep{chow1970optimum,el2010foundations,geifman2017selective,geifman2019selectivenet}. Post-hoc calibration and conformal prediction can support abstention, although they do not shape the representation space during training \citep{guo2017calibration,angelopoulos2021gentle}. Out-of-distribution (OOD) detection methods include maximum softmax probability, ODIN, and Mahalanobis distance, with surveys detailing pitfalls and best practice \citep{hendrycks2017baseline,liang2018enhancing,lee2018simple,mohseni2020self,yang2021generalized}. Energy-based models interpret predictions via an energy landscape in which in-distribution samples receive low energy, and explicit energy training has improved OOD detection; margin-based and contrastive variants further shape the energy function \citep{lecun2006tutorial,grathwohl2019your,liu2020energy}. In RAG, conditioning on retrieved passages improves factuality, yet systems can still over-commit on OOD inputs; prior work explores selective QA, self-evaluation, and retrieval-aware abstention \citep{lewis2020rag,kamath2020selective,kadavath2022language,asai2024selfrag}. Contrastive learning benefits from informative negative mining, widely studied in metric learning and dense retrieval \citep{schroff2015facenet,wu2017sampling,hermans2017defense,karpukhin2020dpr,xiong2021ance}. For medical QA, integrating external corpora helps calibrate a broad not-our-domain boundary; common OOD pools include MedMCQA and SQuAD \citep{rajpurkar2018mchealth,zhang2020medicalsurvey,pal2022medmcqa,rajpurkar2016squad}. Non-parametric scores such as the $k$th-neighbour similarity also provide strong abstention baselines when thresholds are fixed on validation and reused at test \citep{sun2022knn,berthelot2018icarl,hendrycks2017baseline,liang2018enhancing,liu2020energy}. Within this landscape, our approach trains a compact energy-based abstainer directly over dense embeddings, introduces adaptive semi-hard negative mining to expose near-domain confusables, and couples similarity and energy margins, aiming for separation on both clean and hard OOD scenarios in RAG-style QA.

We address this by training an energy-based abstention model that learns a smooth energy landscape over a dense clinical question space. The energy score functions as a calibrated confidence signal: low-energy queries proceed to generation; high-energy queries trigger abstention or escalation. We evaluate on easy and hard abstention splits that include near-distribution confusions which typically defeat probability-based confidence and simple density heuristics.

Our contributions are threefold:
\begin{enumerate}
    \item \textbf{Energy-based abstention for RAG}: We introduce a scalable EBM that enables reliable abstention in dense semantic spaces, outperforming probability and density-based baselines on hard cases despite training with fewer negatives per batch.
    
    \item \textbf{Fair hard-case benchmarking}: We release and evaluate on clean and hard abstention splits under corrected methodology, showing that softmax excels on easy OOD but EBM is more robust on safety-critical hard queries. Softmax’s advantage on easy cases stems from cross-entropy supervision over all positives and negatives in each batch, whereas the EBM trains with only a sampled subset.
    
    \item \textbf{Ablation-driven insights}: We show that energy scoring drives robustness, and that heterogeneous negative exposure (combining hard and easy negatives) is required for generalisation. This provides concrete methodological guidance for designing abstention-aware RAG in safety-critical domains.
\end{enumerate}

\section{Methods}\label{sec:methods}

\subsection{Data Preparation}

We used four main sources of data: (1) \textit{in-domain anchor questions}, derived from a corpus of best-practice clinical guidelines (Supplementary Table~\ref{supp-table-1}) in obstetrics and gynaecology curated by clinicians. Questions were generated from this corpus using ChatGPT-4o, with a subset subsequently validated by the clinical team. From this pool, we selected a representative set of 100K anchors using TF–IDF features and MiniBatchKMeans clustering. (2) \textit{hard negatives}, synthetically generated with a controlled prompt that preserved the original question’s structure and intent (e.g., diagnosis, management, screening) while substituting obstetrics/gynaecology terms with analogues from other specialties (e.g., uterus $\rightarrow$ prostate, CA-125 $\rightarrow$ PSA), thereby producing medically plausible but domain-shifted confounders; (3) \textit{external out-of-domain (OOD) examples}, drawn from publicly available datasets, including the MedMCQA multi-subject medical QA dataset \citep{pal2022medmcqa} and the Stanford Question Answering Dataset (SQuAD) \citep{rajpurkar2016squad}; and (4) a \textit{reserve in-domain corpus} used for mid-range negative sampling. All texts were embedded using the \texttt{BAAI/bge-m3} encoder \citep{xu2023baai_bge} and L2-normalised to ensure consistent cosine similarity across the corpus.

\subsection{Positive and Negative Pair Generation}

For each anchor, the positive was selected using reciprocal-nearest-neighbour (RNN) filtering, retaining a pair only if each was the other’s top-1 neighbour under cosine similarity. This strict reciprocity ensured semantically aligned positives while removing duplicates and one-sided matches \citep{jarvis2010recrnn}. Mid-range negatives were drawn from similarity bands excluded if trivially easy or excessively hard to maintain a balance between informativeness and diversity. Hard negatives were defined as \textit{confusable examples}, i.e., semantically close but clinically misleading text. These were generated using the prompt-based procedure described above and further filtered with the \texttt{google/medgemma-4b-it} model \citep{medgemma2024}, pairing each anchor with one such hard negative to guarantee consistent coverage of confusing samples across splits.

In addition, we constructed an external OOD pool consisting of MedMCQA and SQuAD queries. Sampling from this pool during training encouraged the model to assign higher energy to irrelevant samples, directly improving abstention behaviour. Each training tuple therefore contained an anchor, its RNN-filtered positive, one MedGEMMA-derived hard negative, optional mid-range negatives, and OOD negatives. Since the number of negatives varied across anchors, tuples were padded and masked to enable efficient batch operations.

\subsection{Model Architecture}

Our model adopts a dual-branch design that combines a shared representation space with two task-specific scoring heads. A \textit{projector network} first maps the 1024-dimensional input embeddings (pre-computed using \texttt{BAAI/bge-m3}) into a 256-dimensional latent space. The projector consists of two fully connected layers (1024 $\rightarrow$ 512 $\rightarrow$ 256), with a GELU activation on the hidden layer, followed by L2 normalisation of the output vector. This normalisation ensures stable similarity computations and prevents norm scaling from dominating the training objective.

On top of this shared latent space, we define two alternative heads:
\begin{itemize}
    \item The \textit{energy head}, a two-layer feedforward network (256 $\rightarrow$ 256 $\rightarrow$ 1) with GELU activation, produces a scalar \emph{energy score} for each input. Within the energy-based modelling framework, this score is optimised to separate in-domain anchors and positives from hard negatives, mid-range negatives, and out-of-distribution (OOD) samples.
    \item The \textit{softmax head}, a linear classifier (256 $\rightarrow$ 256 $\rightarrow$ 2) with GELU activation, outputs logits for binary classification between in-domain and out-of-domain classes. This provides a probabilistic baseline against which the abstention-aware energy formulation can be compared.
\end{itemize}

During training, the projector weights are shared across all tuple elements (anchor, positive, and the various negative types). This ensures that all comparisons, whether cosine similarity, energy difference, or classification margin, are carried out in a consistent representation space. Both heads can thus be trained and evaluated fairly on identical embeddings, allowing for a direct comparison of softmax versus energy-based abstention.

\subsection{Semi-Structured Negative Sampling}

Each training mini-batch provides a synchronised tuple $(\mathbf{z}_A,\mathbf{z}_P,\mathbf{z}_{HN})$ for every anchor, where the hard negative $\mathbf{z}_{HN}$ is paired 1--to--1 with its anchor by construction. In addition, we form a heterogeneous candidate pool of negatives per anchor by concatenating: (i) mid-range similarity negatives, (ii) easy negatives, and (iii) uniformly sampled in-domain corpus items. To avoid leakage, these samples exclude items already assigned to training, validation, or test splits.

From this pool, we draw exactly $k_{\text{mine}}$ negatives per anchor \emph{uniformly at random} without replacement. These sampled negatives are used in the loss via a LogSumExp aggregation over per-negative terms. The projector parameters are shared across all tuple elements to ensure that similarity and energy are computed in a single, consistent latent space.

This procedure guarantees (a) a deterministically paired hard negative for each anchor, and (b) a fixed, fair exposure budget of $k_{\text{mine}}$ additional negatives per anchor drawn from a heterogeneous pool.

\subsection{Loss Functions}

\textbf{Energy-Calibrated Semi-Contrastive Triplet Loss (EC-SCTL).} 
For each anchor--positive pair $(\mathbf{z}_A,\mathbf{z}_P)$, we use (i) its deterministically paired hard negative $\mathbf{z}_{HN}$, and (ii) $K$ additional sampled negatives $\mathbf{z}_N$ drawn from the heterogeneous pool. The loss combines similarity and energy terms, drawing on contrastive learning \citep{schroff2015facenet, hermans2017defense}, hard-negative mining \citep{wu2017sampling, karpukhin2020dpr}, and energy-based modelling \citep{grathwohl2019your, liu2020energy}. 

Cosine similarity enforces relative closeness of anchor and positive:
\[
\mathcal{L}_{\text{sim}} = \operatorname{softplus}_T\!\left(m_{\text{sim}} + \cos(\mathbf{z}_A,\mathbf{z}_N) - \cos(\mathbf{z}_A,\mathbf{z}_P)\right),
\]
where $\operatorname{softplus}_T(x) = T \cdot \log\!\big(1 + \exp(x/T)\big)$ with temperature $T=\texttt{softplus\_temp}$. 
This smooths margin violations, avoiding sharp hinge cut-offs.

The energy term encourages positives to have lower energy than negatives:
\[
\mathcal{L}_{\text{energy}} = \lambda \cdot \operatorname{softplus}_T(E_P - E_N + m_E).
\]

The $K$ sampled negatives are aggregated with a LogSumExp:
\[
\mathcal{L}_{\text{core}} = \tfrac{1}{T} \log \sum_{k=1}^K \exp \big( T (\mathcal{L}_{\text{sim},k} + \mathcal{L}_{\text{energy},k}) \big),
\]
where $T$ interpolates between mean pooling ($T \to 0$) and max pooling ($T \to \infty$). 
This lets the model adapt between averaging across all negatives and focusing on the hardest ones.

Two auxiliary hinge terms stabilise training: (i) an OOD hinge against the hardest external negative, $\mathcal{L}_{\text{OOD}}$, and (ii) the paired hard-negative hinge using $\mathbf{z}_{HN}$, $\mathcal{L}_{\text{HN}}$. 
The final objective is:
\[
\mathcal{L}_{\text{EC-SCTL}} = \mathcal{L}_{\text{core}} + w_{\text{OOD}}\mathcal{L}_{\text{OOD}} + w_{\text{HN}}\mathcal{L}_{\text{HN}}.
\]

\textbf{Softmax baseline loss.} 
The baseline classifier is trained with cross-entropy between in-domain ($y=0$) and OOD ($y=1$) classes:
\[
\mathcal{L}_{\text{softmax}} = - \sum_{c \in \{0,1\}} y_c \log p_c,
\quad p_c = \operatorname{softmax}(\mathbf{z})_c.
\]

\subsection{Training and Evaluation Strategy}

All models were trained under a consistent protocol to ensure fair comparison. 
For both the energy-based and softmax models, input embeddings were mapped into a shared 256-dimensional latent space via the projector network. 
Optimisation used AdamW (learning rate $10^{-3}$, weight decay $10^{-4}$) with cosine annealing over 20 epochs and batch size 1024. 
Each anchor was paired with its positive and a synchronised hard negative, and further supplemented with exactly $k_{\text{mine}}$ additional negatives. 
This ensured that both the energy-based and softmax heads were exposed to the same number of negatives per batch. 
The KNN baseline was built on the same pre-computed BGE-M3 embeddings, indexed with FAISS \citep{johnson2019billion} for cosine similarity, using $k=5$ neighbours.

Validation loss was monitored throughout training, and the checkpoint with the lowest value was retained. 
Abstention thresholds were calibrated on the validation split and then fixed for evaluation on the test set. 
We report results at two operating points:  
(i) the detection error threshold $\tau_{\text{DetErr}}$, defined as
\[
\text{DetErr} = \tfrac{1}{2}\big(\text{FPR}(\tau) + \text{FNR}(\tau)\big),
\]
which minimises the average of false positive and false negative rates; and  
(ii) the $\tau_{95}$ threshold, corresponding to the operating point closest to 95\% TPR.

Evaluation covered two scenarios: (a) {ID vs.~clean OOD}, comparing in-domain anchors against an external OOD pool balanced to ID size, and (b) {ID vs.~hard OOD}, comparing anchors against their paired synthetic hard negatives. 
For each method we report AUROC, AUPRC, the false positive rate at $\tau_{95}$ (FPR@95), and the detection error (DetErr).

To assess design choices, we conducted ablations on the energy model: removing the energy head (\texttt{no\_energy}) to isolate the similarity-only loss, and removing the external OOD hinge (\texttt{no\_ext\_ood}) to test its contribution beyond paired hard negatives. 
Implementation details (mixed precision, gradient clipping, hardware) are provided in the Appendix.

\begin{table*}[htbp]
\centering
\scriptsize
\setlength{\tabcolsep}{2pt}
\caption{Comparison of EBM, Softmax, and KNN across training configurations.}
\label{tab:full_results_compact}
\begin{tabular}{p{1.2cm}lccc}
\toprule
Config & Method & 
\makecell{Hard \\ \footnotesize AUROC/FPR@95 \\ \footnotesize AUPR, DetErr} &
\makecell{Easy \\ \footnotesize AUROC/FPR@95 \\ \footnotesize AUPR, DetErr} &
\makecell{Mixed \\ \footnotesize AUROC/FPR@95 \\ \footnotesize AUPR, DetErr} \\

\midrule
\multirow{3}{*}{\makecell{All \\ Negs}} 
  & EBM     & [0.961 / 0.265], [0.973, 0.073] & [0.981 / 0.036], [0.988, 0.042] & [0.970 / 0.123], [0.980, 0.058] \\
  & Softmax & [0.949 / 0.358], [0.961, 0.102] & [0.995 / 0.006], [0.996, 0.021] & [0.971 / 0.169], [0.978, 0.067] \\
  & KNN     & [0.855 / 0.006], [0.832, 0.429] & [0.998 / 0.006], [0.999, 0.015] & [0.926 / 0.006], [0.936, 0.221] \\
\midrule
\multirow{3}{*}{\makecell{Hard \\ Only}} 
  & EBM     & [0.964 / 0.251], [0.976, 0.061] & [0.720 / 0.971], [0.772, 0.302] & [0.845 / 0.890], [0.891, 0.194] \\
  & Softmax & [0.955 / 0.325], [0.967, 0.090] & [0.765 / 0.978], [0.814, 0.259] & [0.861 / 0.896], [0.901, 0.181] \\
  & KNN     & [0.855 / 0.006], [0.832, 0.429] & [0.998 / 0.006], [0.999, 0.015] & [0.926 / 0.006], [0.936, 0.221] \\
\midrule
\multirow{3}{*}{\makecell{Easy \\ Only}} 
  & EBM     & [0.512 / 0.983], [0.561, 0.456] & [0.997 / 0.000], [0.998, 0.009] & [0.753 / 0.946], [0.830, 0.243] \\
  & Softmax & [0.577 / 0.958], [0.595, 0.428] & [0.997 / 0.001], [0.998, 0.010] & [0.787 / 0.911], [0.848, 0.243] \\
  & KNN     & [0.855 / 0.006], [0.832, 0.429] & [0.998 / 0.006], [0.999, 0.015] & [0.926 / 0.006], [0.936, 0.221] \\
\midrule
\multirow{3}{*}{\makecell{No \\ Hard}} 
  & EBM     & [0.507 / 0.967], [0.544, 0.469] & [0.997 / 0.000], [0.998, 0.009] & [0.750 / 0.925], [0.826, 0.245] \\
  & Softmax & [0.586 / 0.958], [0.606, 0.416] & [0.997 / 0.001], [0.998, 0.011] & [0.790 / 0.914], [0.850, 0.240] \\
  & KNN     & [0.855 / 0.006], [0.832, 0.429] & [0.998 / 0.006], [0.999, 0.015] & [0.926 / 0.006], [0.936, 0.221] \\
\midrule
\multirow{3}{*}{\makecell{No \\ Easy}} 
  & EBM     & [0.964 / 0.263], [0.976, 0.060] & [0.633 / 0.996], [0.725, 0.325] & [0.803 / 0.980], [0.870, 0.200] \\
  & Softmax & [0.954 / 0.336], [0.967, 0.089] & [0.663 / 0.999], [0.753, 0.304] & [0.812 / 0.992], [0.874, 0.200] \\
  & KNN     & [0.855 / 0.006], [0.832, 0.429] & [0.998 / 0.006], [0.999, 0.015] & [0.926 / 0.006], [0.936, 0.221] \\
\midrule
\multirow{3}{*}{\makecell{Hard \\ + Easy}} 
  & EBM     & [0.960 / 0.299], [0.973, 0.072] & [0.983 / 0.030], [0.989, 0.039] & [0.970 / 0.122], [0.980, 0.056] \\
  & Softmax & [0.948 / 0.347], [0.960, 0.107] & [0.996 / 0.006], [0.997, 0.020] & [0.971 / 0.169], [0.978, 0.068] \\
  & KNN     & [0.855 / 0.006], [0.832, 0.429] & [0.998 / 0.006], [0.999, 0.015] & [0.926 / 0.006], [0.936, 0.221] \\
\bottomrule
\end{tabular}
\end{table*}

\section{Results}\label{sec:results}

Table~\ref{tab:full_results_compact} reports AUROC, AUPR, and calibration metrics (FPR@95 and DetErr) across training configurations and evaluation conditions (ID vs.~hard OOD, easy OOD, and mixed). Three main findings emerge.
Whenever training included hard negatives (\texttt{all\_negs}, \texttt{hard\_only}, \texttt{hard+easy}), EBM achieved higher AUROC and lower DetErr than softmax on the hard-OOD split. 
For example, in the full-data setting (\texttt{all\_negs}), AUROC improved from 0.949 (Softmax) to 0.961 (EBM) and DetErr dropped from 0.102 to 0.073, a $\sim$30\% relative reduction. 
This shows that shaping the energy landscape provides a more reliable abstention signal for near-domain confusables.
\begin{figure*}[ht]
    \centering
    \includegraphics[width=0.7\linewidth]{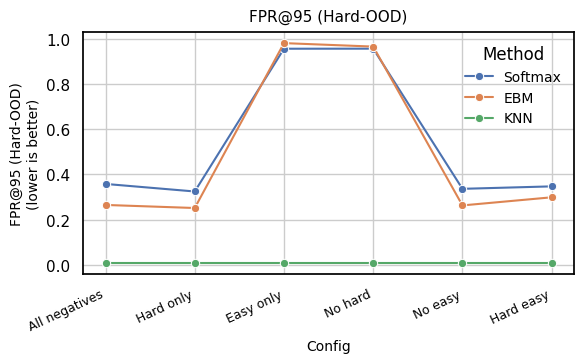}
    \caption{FPR@95 on the hard-OOD test split for models trained with different negative sampling configurations.}

    \label{fig:ood_fpr}
    
    \vspace{1em} 

    \includegraphics[width=0.7\linewidth]{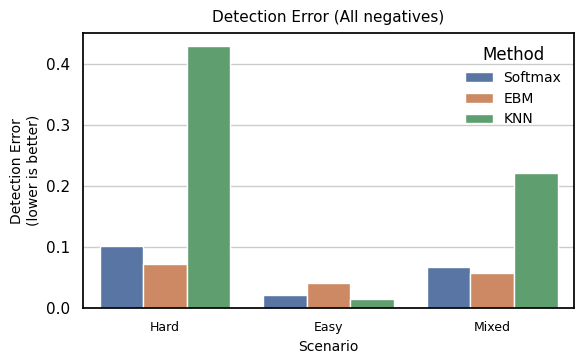}
    \caption{Detection Error on the test split (All negatives training configuration), comparing EBM, Softmax, and KNN.}

    \label{fig:ood_deterr}
\end{figure*}

When trained only on hard negatives (\texttt{hard\_only}), both models failed to generalise to clean OOD: AUROC fell below 0.80 and FPR@95 rose to $\sim$97--98\%. 
Including easy negatives (\texttt{all\_negs} or \texttt{hard+easy}) restored strong performance, with AUROC above 0.98 and FPR@95 below 0.04. 
This highlights that heterogeneous negative exposure, not just hard cases, is required for robust decision boundaries.

On easy-OOD, $k$-NN achieved near-perfect AUROC ($>0.998$) with very low DetErr (0.015), outperforming both parametric models. 
On mixed-OOD, it reached AUROC 0.926, showing reasonable robustness. 
However, on hard-OOD its AUROC dropped to 0.855 and DetErr exceeded 0.42, indicating that simple nearest-neighbour density fails against confusable examples where EBM retains clear margins.

Removing hard negatives (\texttt{no\_hard}, \texttt{easy\_only}) or easy negatives (\texttt{no\_easy}) sharply degraded mixed-OOD performance. 
For instance, AUROC dropped from 0.970 (\texttt{all\_negs}) to 0.787 (\texttt{easy\_only}) and 0.803 (\texttt{no\_easy}). 
This validates our design choice of pairing each anchor with a synchronised hard negative while sampling additional easy and mid-range negatives.

Overall, the results show that EBM consistently improves abstention on safety-critical hard-OOD queries, while softmax and $k$-NN are sufficient only for clean OOD rejection. Robust abstention requires both energy calibration and balanced negative exposure.

\section{Discussion}\label{sec:discussion}

This study demonstrates that energy-based modelling offers a more reliable abstention mechanism than probability- or density-based approaches, particularly on semantically hard, near-distribution queries. By shaping the latent representation space to enforce separation between in-domain anchors and confusable negatives, our method reduces over-commitment on queries that are most likely to elicit persuasive but unsafe responses. These findings position abstention not as an auxiliary safeguard, but as a necessary foundation for trustworthy retrieval-augmented generation (RAG). In safety-critical domains, a system that cannot reliably defer when evidence is insufficient risks undermining both clinical outcomes and user trust. An additional challenge arises when queries fall within the broad medical domain yet outside the scope of the indexed corpus. For example, a RAG system trained on obstetrics and gynaecology guidelines may still receive questions from related but unrepresented areas, such as oncology or paediatric gynaecology. In such cases, generating an answer risks conveying unwarranted authority without evidential grounding. The ability to abstain from these near-domain but out-of-corpus queries is therefore as important as rejecting irrelevant prompts altogether, ensuring that the system only responds when retrieval can provide coverage within its validated evidence base.
\newline

Our findings highlight three broader implications. First, shaping the energy landscape provides a fundamentally different abstention signal than softmax probabilities: rather than relying on probability calibration, the energy-based approach enforces structured separation that remains robust when queries are semantically close to the training distribution. Second, the role of heterogeneous negatives is methodological as well as practical. Exposure to both easy and hard negatives is required for the model to learn not only to reject trivially irrelevant queries but also to distinguish clinically misleading near-distribution confusions, which are especially hazardous in safety-critical domains. Third, the comparison with $k$NN illustrates the limits of non-parametric density estimation. While such methods perform strongly on clean out-of-distribution (OOD) cases, they fail when subtle semantic confusions arise, underscoring the value of parametric shaping of the latent space. From an applied perspective, these results suggest that abstention-aware models should be designed with explicit energy calibration, structured negative exposure, and evaluation protocols that include both clean and confusable OOD cases. This design pattern is particularly relevant for medical QA, where over-confident errors can directly affect patient safety.\newline

The importance of robust abstention is especially clear in women’s health. Obstetrics and gynaecology involve high-stakes, time-sensitive decisions where overconfident errors may directly affect maternal or neonatal outcomes. Large language models such as ChatGPT have shown promise but also serious limitations in this setting. For example, evaluation against the Royal College of Obstetricians and Gynaecologists (RCOG) MRCOG examination revealed only moderate accuracy on basic science and coin-toss performance on clinical reasoning, with the model frequently expressing high confidence in incorrect answers \citep{Bachmann2024WomenHealthLLM}. This overconfidence represents a substantive risk if left unmitigated. Subsequent work introduced \emph{semantic entropy} as a meaning-level uncertainty signal, which outperformed perplexity for detecting unreliable outputs on obstetrics and gynaecology questions \citep{PennyDimri2025SemanticEntropy}. Expert validation showed that semantic entropy could discriminate between correct and incorrect responses even when surface fluency was indistinguishable. However, this approach requires generating multiple candidate answers to estimate the distribution of meanings, incurring both latency and computational overhead.\newline

Our results should be seen as complementary to these prior contributions. Energy-based abstention offers an efficient, pre-generation mechanism, while semantic entropy provides a high-fidelity, meaning-aware signal post-generation. A layered strategy is therefore possible: EBMs could serve as a fast first-line filter, with semantic entropy applied selectively to borderline cases or in high-stakes scenarios where additional certainty is required. Women’s health represents an exemplar domain for this combined approach, as its diagnostic and treatment gaps heighten the need for both efficiency and robustness. In low-resource or inequitable contexts, where escalation pathways to specialist expertise may be limited, abstention mechanisms that can act rapidly and scale reliably are especially valuable.\newline

In comparison with existing approaches to abstention and model safety, EBMs confer several advantages. Softmax probabilities provide only a proxy for confidence and are brittle under distributional shift. Perplexity-based thresholds require token-level likelihoods, which do not capture semantic misalignment. Semantic entropy improves discrimination but is computationally intensive. Retrieval-aware methods and rule-based safeguards can provide useful signals, yet they rely on additional prompts or handcrafted heuristics that are difficult to generalise. By contrast, the EBM operates pre-generation, requires no multiple decoding passes, scales to large corpora, and demonstrates robustness in handling near-domain confusions where probability-based methods fail.\newline

From a methodological perspective, this work makes several contributions. We introduce an energy-based scoring head trained over dense embeddings, showing that latent-space shaping yields more reliable abstention than probability calibration. We develop a semi-structured negative sampling strategy that balances easy, mid-range, and hard negatives, ensuring robust decision boundaries. We also correct benchmarking practice by explicitly separating clean and hard out-of-distribution cases, exposing performance differences that standard splits obscure. These advances provide a principled framework for abstention in retrieval-augmented models. For medicine, the work delivers a practical abstention mechanism for women’s health RAG, reducing unsafe overconfidence and supporting the safe integration of AI into clinical workflows. For computer science, it establishes an energy-based abstention method that outperforms perplexity, semantic entropy, and softmax baselines, while offering ablation-driven insights into how negative exposure sharpens decision boundaries.\newline

There are, however, important limitations. The dataset is restricted to English guideline-derived corpora, limiting immediate applicability to multilingual settings. Synthetic hard negatives, though clinically plausible, may not capture the full complexity of real-world queries. No prospective evaluation with clinicians-in-the-loop has yet been conducted, leaving the impact of abstention on real-world decision-making untested. Architecturally, the abstention mechanism operates at the embedding level and has not been fully integrated into an end-to-end generative pipeline. Finally, threshold calibration is static and may need adaptation for deployment under varying conditions.\newline

Future work should extend this approach to multilingual corpora to improve global equity of access. Adaptive semi-hard negative mining could refine decision boundaries further. Prospective evaluation in clinical workflows will be critical to assess whether abstention supports safe escalation and trust. Hybrid systems that combine the efficiency of EBMs with the discriminative power of semantic entropy could deliver both scalability and high-fidelity safety checks. Engagement with regulatory frameworks such as the EU AI Act and FDA guidance will also be essential to formalise abstention as a safety mechanism within clinical AI systems.\newline

Reliable abstention is a first-order requirement for safe AI in medicine. This study shows that shaping the latent energy landscape provides an efficient and robust mechanism for rejecting unsafe queries. Women’s health, with its persistent diagnostic and treatment gaps, is a critical test-bed for such approaches. By combining fast, energy-based abstention with discriminative semantic entropy, future RAG systems can move towards equitable, trustworthy clinical support, reducing the risk of overconfident error while enabling scalable deployment.

\section{Conclusion}\label{sec:conclusion}

We presented an energy-based abstention framework for retrieval-augmented question answering that combines similarity and energy margins in a shared latent space. Our approach consistently improves robustness on hard OOD cases while remaining competitive on clean OOD detection. Beyond performance, the work establishes a design principle: reliable abstention emerges when representation learning and scoring are trained jointly under heterogeneous negative exposure. Future research should extend this approach to multilingual and adaptive settings, and explore its integration with end-to-end generative models for safe RAG systems.

\section{Funding}

This research was supported by the UKRI Medical Research Council (MR/X029689/1).

\section{Author Contributions}

RS contributed to conceptualisation, methodology design, formal analysis, and drafting of the manuscript.  GDJ contributed to conceptualisation, methodology, software development, supervision, manuscript review and editing, and secured funding.  SW and LL contributed to methodology and manuscript review and editing.  MB and AS contributed to data curation.  BA contributed to project administration.

\section{Competing Interests}

The authors have no conflicts of interest to declare.

\bibliography{sn-article}

\clearpage
\appendix
\section{Appendix : Supplementary Table}

\footnotesize
\renewcommand{\arraystretch}{0.9} 
\setlength{\tabcolsep}{3pt}       

\begin{longtable}{p{2.8cm} p{4cm} p{0.8cm} p{2.2cm} p{2cm}}
\caption{An excerpt of women's health guidelines and publications used to prepare the questions corpus in this research. Note: The full list has been submitted as the supplementary material.} \\
\label{supp-table-1} \\
\toprule
\textbf{Title} & \textbf{Synopsis} & \textbf{Year} & \textbf{Authors} & \textbf{Publisher} \\
\midrule
\endfirsthead

\multicolumn{5}{c}%
{{\bfseries \tablename\ \thetable{} -- continued from previous page}} \\
\toprule
\textbf{Title} & \textbf{Synopsis} & \textbf{Year} & \textbf{Authors} & \textbf{Publisher} \\
\midrule
\endhead

\midrule
\multicolumn{5}{r}{{Continued on next page}} \\
\midrule
\endfoot

\bottomrule
\endlastfoot

Workplace-based assessment: a new approach to existing tools 
& The article discusses the implementation challenges and revisions of workplace-based assessment (WPBA) tools in obstetrics and gynaecology, emphasizing the distinction between formative and summative assessments. 
& 2014 
& William Parry-Smith, Ayesha Mahmud, Alex Landau, et al. 
& TOG \\

Contraceptive methods and issues around the menopause: an evidence update 
& The publication discusses recent advances in contraceptive methods available to perimenopausal women, issues related to menopause, and the integration of hormone replacement therapy with contraception. 
& 2017 
& Shagaf H Bakour, Archana Hatti, Susan Whalen 
& TOG \\

Twin and triplet pregnancy 
& This guideline covers care for pregnant women and pregnant people with a twin or triplet pregnancy, aiming to reduce complications and improve outcomes. 
& 2019 
& -- 
& NICE \\

Management of sickle cell disease in pregnancy. A British Society for Haematology Guideline 
& This guideline describes the management of sickle cell disease in pregnancy, covering preconception, antenatal, intrapartum, and postnatal care, with updates on genetic diagnosis, medication review, and antenatal care recommendations. 
& 2021 
& Eugene Oteng-Ntim, Sue Pavord, Richard Howard, et al. 
& Br J Haematol \\

Inducing labour 
& This guideline covers the circumstances for inducing labour, methods of induction, assessment, monitoring, pain relief, and managing complications to improve advice and care for pregnant women considering or undergoing induction of labour. 
& 2021 
& -- 
& NICE \\

Saving Lives, Improving Mothers' Care State of the Nation Report 
& The report presents surveillance findings and lessons learned to inform maternity care from the UK and Ireland Confidential Enquiries into Maternal Deaths, focusing on thrombosis, thromboembolism, malignancy, ectopic pregnancy, and the care of recent migrants with language difficulties from 2020-2022. 
& 2024 
& MBRRACE-UK 
& NPEU, Univ. Oxford \\

Laparoscopy in urogynaecology 
& This article discusses the advancements and challenges in laparoscopic urogynaecological surgery, focusing on procedures for prolapse and stress incontinence, and highlights the importance of training and patient choice. 
& 2018 
& Rajvinder Khasriya, Arvind Vashisht, Alfred Cutner 
& TOG \\

Failed hysteroscopy and further management strategies 
& This article explores various methods to overcome cervical stenosis in hysteroscopy, highlighting techniques such as pharmacological, mechanical, hygroscopic, and ultrasound-guided dilatation. 
& 2016 
& Sophie Relph, Tessa Lawton, Mark Broadbent, et al. 
& TOG \\

An overview of assisted reproductive technology procedures 
& This article provides a comprehensive overview of assisted reproductive technology (ART) procedures, discussing key steps, individualised treatment protocols, and multidisciplinary management for fertility issues. 
& 2018 
& Harish M Bhandari, Meenakshi K Choudhary, Jane A Stewart 
& TOG \\

Multiple Pregnancies Following Assisted Conception 
& This paper discusses the high rates of multiple pregnancies following IVF and the measures taken to reduce these rates, highlighting the importance of elective single embryo transfer (eSET) and the progress made in the UK. 
& 2018 
& T El-Toukhy, S Bhattacharya, VA Akande 
& RCOG \\

Making Abortion Safe 
& The publication outlines best practices in abortion care to ensure safe and effective services, emphasizing evidence-based clinical practice and addressing barriers to safe abortion globally. 
& 2022 
& Updated by Sharon Cameron, Jayne Kavanagh, Patricia A Lohr; originally developed by Anna Glasier and others. 
& RCOG \\

Evaluating misoprostol and mechanical methods for induction of labour 
& This paper reviews the use of misoprostol and mechanical methods for induction of labour, comparing their safety and efficacy to the standard drug, dinoprostone. 
& 2022 
& Andrew D. Weeks, Kate Lightly, Ben W. Mol, et al. 
& RCOG \\

Does ovarian cystectomy pose a risk to ovarian reserve and fertility? 
& This article reviews the impact of ovarian cystectomy on ovarian reserve and fertility, discussing factors such as cyst type, size, and surgical techniques that influence outcomes. 
& 2020 
& Neerujah Balachandren, Ephia Yasmin, Dimitrios Mavrelos, et al. 
& TOG \\

\end{longtable}

\end{document}